\documentclass{article}
\usepackage{spconf,amsmath,graphicx,amssymb,subcaption}

\usepackage{graphicx}
\usepackage{multirow}
\usepackage{multicol}

\title{XNOR-former: Learning Accurate Approximations in Long Speech Transformers }
\name{
Roshan Sharma$^1$, Bhiksha Raj$^2$}
\address{$^1$Electrical and Computer Engineering,$^2$Language Technologies Institute \\ Carnegie Mellon University 
}
\begin{document}
\ninept
\maketitle
\begin{abstract}
Transformers are among the state of the art for many tasks in speech, vision, and natural language processing, among others. Self-attentions, which are crucial contributors to this performance have quadratic computational complexity, which makes training on longer input sequences challenging. Prior work has produced state-of-the-art transformer variants with linear attention, however, current models sacrifice performance to achieve efficient implementations. In this work, we develop a novel linear transformer by examining the properties of the key-query product within self-attentions. Our model outperforms state of the art approaches on speech recognition and speech summarization, resulting in 1 \% absolute WER improvement on the Librispeech-100 speech recognition benchmark and a new INTERVIEW speech recognition benchmark, and 5 points on ROUGE for summarization with How2.
\end{abstract}
\section{Introduction}
\label{sec:intro}
Transformers are among the state of the art for many tasks in speech, natural language processing and vision, among other fields involving ordered sequences of data. They achieve their exceptional generalization in part due to ``multi-head self-attention'' -- processing blocks which derive (a set of) updated representations for each input in a sequence as a weighted sum of values derived from all the inputs in the sequence. This computation, however, has a computational and space complexity that is quadratic in the length of the input sequence. 
%
As a result, it becomes challenging to operate on longer input sequences on modern GPUs and TPUs, and training such models from scratch is time and labor intensive. 




The computational challenges arise from the formulation of self-attention. For an input sequence of length $N$ each head of the multi-head self attention block derives three $N$-row matrices: a \textit{value} matrix $V$ representing the latent-representation vectors for the $N$ inputs in the sequence, a \textit{query} matrix $Q$ representing the probes with which each input derives the ``self-attention'' weights required to update itself, and a \textit{key} matrix $K$, representing the key contribution of each input to the computation of its own weight in updating any input. The actual updated representations for the input computed by the head has the form $softmax(QK^\top)V$.  The bottleneck arises from $softmax(QK^\top)$: both $Q$ and $K$ have $N$ rows, $QK^\top$ requires $O(N^2)$ computation. This cannot be factored, since the softmax operates on the \emph{product}, which must necessarily be computed before the softmax is applied.

Recently, many works have had success in developing attention mechanisms that approximate the standard self-attention in linear time and space complexity.

Pattern based methods \cite{dai-etal-2019-transformer,Beltagy2020Longformer,Zaheer2020BigBird,Rae2019CompressiveTransformer} save on computation by limiting the size of the input blocks they can work on, however they potentially lose the larger context of the input in the process. Sparsity based methods \cite{xiong2021nystromformer,wang2020linformer} use low-rank approximations of attention to reduce compuation; however they are, by nature lossy, approaching the computational complexity of full self-attention if losslessness is to be assured.  Linear transformers\cite{zhen2022cosformer,katharopoulos_et_al_2020,sun2022} do achieve linear-time and can thus operate on long sequences. However, current linear transformers still have significant performance degradations compared to the standard self-attention.

In this paper, we attempt to bridge this gap by developing the \emph{XNOR self-attention} and correspondingly, the \emph{XNOR-former}. 
Our work falls into the pool of work on linear transformers that use the kernel trick to compute self-attentions in linear time.  We observe that $\sigma(pq) \approx \sigma(p) \bar{\oplus} \sigma(q)$, where $\sigma()$ is the sigmoid function, and $\bar{\oplus}$ is the $XNOR$ operator for Boolean variables, but more generally can be expressed as $x \bar{\oplus} y = xy + (1-x)(1-y),~~x,y \in [0,1]$. This enables us to decompose the softmax over the product $QK^\top$ in terms of the $XNOR$ of the softmax over the individual terms $Q$ and $K^\top$. This reduces the computation of the attention, and in fact the overall self-attention update, to linear time.




Also, native transformer self-attentions are position agnostic, and it has been found advantageous to encode positional information within self-attention\cite{zhen2022cosformer,su2021roformer,dai-etal-2019-transformer} through positional embeddings. However standard mechanisms are unsuitable for linear transformers since they require the explicit computation of attention weights. Inspired by \cite{zhen2022cosformer,su2021roformer}, we explore the use of rotational and cosine relative positional embeddings and integrate them with the proposed XNOR-former.

We evaluate our proposed approach on Speech recognition and summarization tasks -- tasks which are characterized by very long input sequences (typically 10-100x longer than the image- and text-based inputs generally associated with transformers). 
Our proposed approach outperforms prior kernel methods on two speech recognition benchmarks and end-to-end speech summarization, while maintaining linear-in-length computational and memory expense.

The rest of this paper is organized as follows: Section \ref{sec:proposed_approach} describes the proposed approach, Section \ref{sec:setup} explains design and evaluation of our experiments, Section \ref{sec:results} details our experimental results and ablation studies. 

\section{X-NOR Self Attention}
\label{sec:approach}

\label{sec:proposed_approach}
\subsection{Standard Self-Attention}
Consider an input sequence $X = [X_1; X_2;  \cdots; X_N]$ (representing a stacking of $N$ row vectors) of length $N$ to a multi-head self-attention module. This module comprises several self-attention ``heads''.  A self-attention head \cite{vaswani2017} computes, from the input $X$, a key matrix $K=XW_K \in  \mathbb{R}^{N \times d}$, a query matrix  $Q=XW_Q \in  \mathbb{R}^{N \times d}$, and a value matrix $V=XW_V \in  \mathbb{R}^{N \times d}$, where $W_K,~W_Q$ and $W_V$ are its parameters. The output of this self-attention $\mathcal{O} \in \mathbb{R}^{N \times d} = [ \mathcal{O}_1 ; \mathcal{O}_2 ; \cdots  ; \mathcal{O}_N ]$ is then computed as
\begin{equation}
    \label{equation:vaswani_attn}
    \mathcal{O} = Sm\Big(\frac{QK^\top}{\sqrt{d}}\Big) V
\end{equation}
where $Sm()$ represents the softmax operator, applied row-wise to its matrix argument. The computational complexity of computing $QK^\top$ (and hence $\mathcal{O}$) is quadratic in $N$. This is not viable to compute on long input sequences.

To facilitate simplification of the computation of self attention, the softmax may be replaced by a normalized similarity measure:
\begin{equation}
    \mathcal{O}_i =  \frac{ \sum_{j=1}^N S(Q_i,K_j)}{\sum_{j} S(Q_i,K_j)}.V_j
\label{eq:smsim}
\end{equation}
where $S(Q_i,K_j)$ is the similarity between the $i^{\rm th}$ query and the $j^{\rm th}$ key. In fact, the softmax is a special case of the above where $S(Q_i,K_j) = \exp(Q_iK_j^\top)$. 

\subsection{Linear Transformers}
The $O(N^2)$ complexity of softmax-based attention results from inner-product term $Q_iK_j^\top$ that occurs within the similarity term $S(Q_i,K_j) \sim \exp(Q_iK_j^\top)$ for the softmax and prevents factorization of the computation.  Linear transformers replace the exponent over inner products by a factored ``product over Kernels'' similarity measure:  $S(\mathcal{Q},\mathcal{K}) = \phi(\mathcal{Q})\phi(\mathcal{K})^\top$, where $\phi()$ is a ``Kernel'' function. Using the associative law of matrix multiplication, this permits the computation to be of the following form, which can be computed in linear time.
\begin{equation}
    \label{equation:general_linear_attn}
    \mathcal{O}_i = \sum_{j=1}^N \frac{\phi(Q_i)(\phi(K_j)V_j)}{\sum_{j} \phi(Q_i)\phi(K_j)}.
\end{equation}
Prior works have used Gaussian Kernels\cite{choromanski2021rethinking}, ELU based kernels\cite{katharopoulos_et_al_2020}, and ReLU kernels\cite{zhen2022cosformer}. The closest to our proposed approach is the softmax Kernel, which uses $\phi() = Sm()$. Though these succeed in reducing the computational overhead, they often result in undesirable performance reductions, compared to the full softmax-based attention.



\subsection{Examining the Sigmoid of the Product}
Our objective is to close the gap between linear transformers and the full transformer formulation without losing the computational advantage of linear time. As mentioned earlier, the main impediment to linear-time factorization is the softmax over the product in the conventional transformer.  

\begin{figure}[t]
    \centering
    \begin{subfigure}{0.2395\textwidth}
        \includegraphics[width=1.0\textwidth]{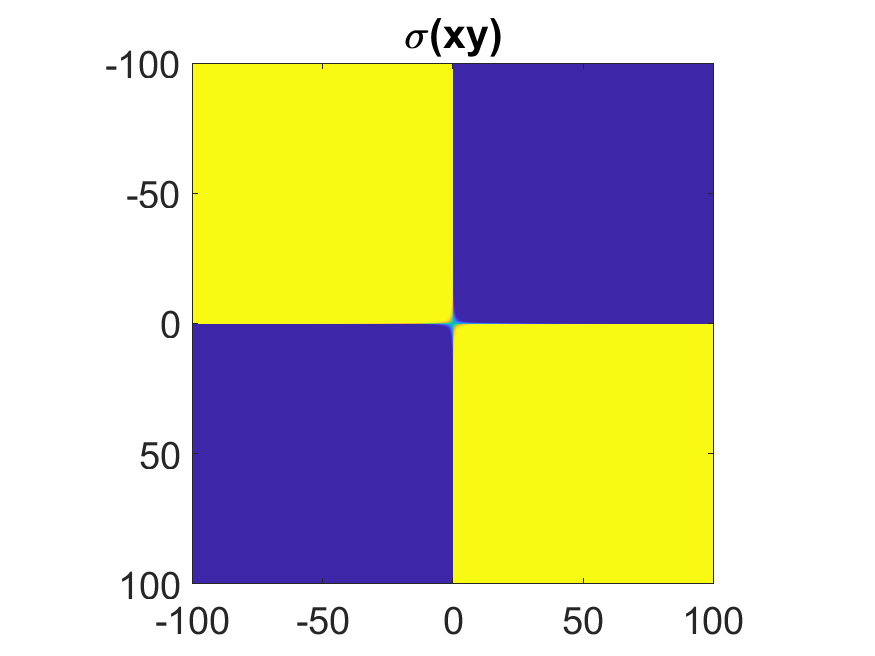}
    \end{subfigure}
    \begin{subfigure}{0.2395\textwidth}
        \includegraphics[width=1.0\textwidth]{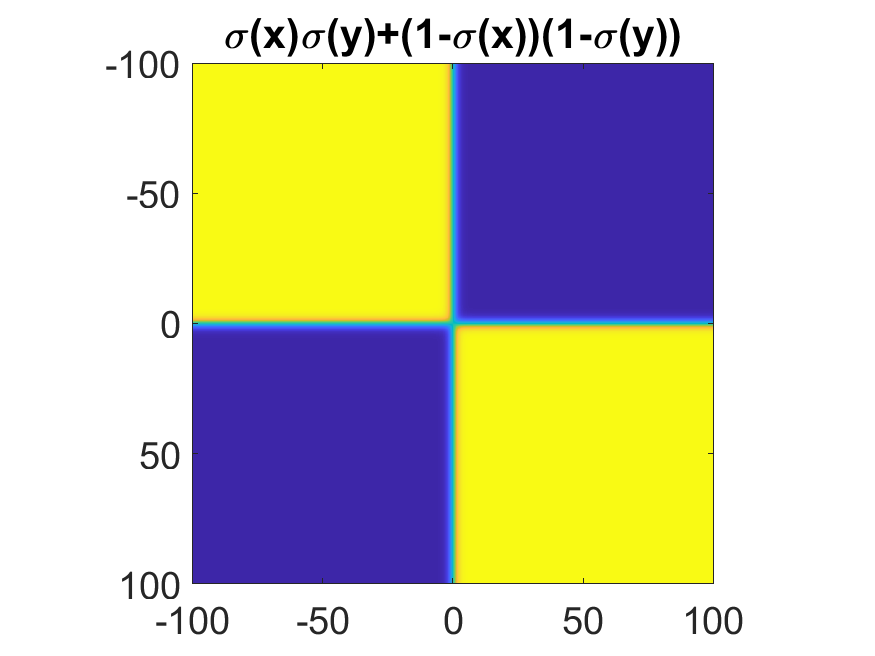}
    \end{subfigure}
    \caption{(a) Left: $\sigma(xy)$ for $x,y \in (-100,100)$. (b) Right: The approximation $\sigma(x)\bar{\oplus}\sigma(y)$.}
    \label{fig:sigmoid_xy_plot}        
\end{figure}

Our proposal builds on the following insight: consider the simplest version of a softmax, which is the sigmoid. $\sigma(xy)$, the sigmoid of the product of two variables $x$ and $y$, has a distinct XNOR-like ($\bar{oplus}$-like) behavior as shown in Figure \ref{fig:sigmoid_xy_plot} a.  This cannot be factored because the XNOR cannot be modelled by a linear boundary. However, analogously to the decomposition of the XNOR, $\sigma(xy)$ \textit{can} be expressed as the sum of \textit{two} bilinear terms: 
\begin{equation}
\label{equation:sigmoid_xnor_factorization}
\sigma(xy) \approx \sigma(x) \bar{\oplus} \sigma(y) = \sigma(x)\sigma(y) + (1-\sigma(x))(1-\sigma(y))
\end{equation}
Figure \ref{fig:sigmoid_xy_plot}b shows the approximation $\sigma(x)\bar{\oplus}\sigma(y)$. Except for a narrow region near the axes, the error is minimal.

This leads us to our formulation for the XNOR-former.

%


\subsection{X-NOR Self-Attention}
\label{sec:xnor_selfattn}
Drawing from the X-NOR approximation of the sigmoid, we  now extend this to approximate the softmax product in Equation \ref{eq:smsim} as
\begin{equation}
    \label{equation:softmax_xnor_factorization}
    S(Q_i,K_j) = w_1Sm(Q_i)Sm(K_j)+ w_2Sm'(Q_i)Sm'(K_j),
\end{equation}
where, as before, $Sm$ represents the softmax operator and $Sm' = 1- Sm$ represents its complement.  The weights $w_1$ and $w_2$ account for the fact that the variable is of length $N$ as opposed to binary.

This gives us the following XNOR factorization of self-attention, which allows it to be computed in linear time and space complexity. 
%
\begin{equation}
\mathcal{O}_i  = \frac{ \sum_{j=1}^N w_1 Sm(Q_i)(Sm(Kq_j)V_j) + w_2 Sm'(Q_i)(Sm'(K_j)V_j)}{\sum_j w_1 Sm(Q_i)Sm(K_j) + w_2 Sm'(Q_i)Sm'(K_j)}
    \label{equation:softmax_wxnor}
\end{equation}
For  simple XNOR self-attention $w_1 = w_2 = 1$. We also make the weights learnable. We refer to this variant as the weighted-XNOR or W-XNOR self-attention.

\subsection{Positional Encodings}
Equation \ref{equation:softmax_wxnor}, which captures feature-based attention weights, however does not consider the positional relation of the inputs in a sequence -- reordering inputs will not change their weights. 

To remedy this, we require \textit{positional embeddings} within the attention mechanism to incorporate awareness of the relative position $i-j$ or absolute positions $i,j$ within the sequence. Relative positional embeddings were demonstrated to outperform absolute embeddings in \cite{dai-etal-2019-transformer,shaw2018self}. Learnable relative positional encodings, introduced in \cite{dai-etal-2019-transformer} rely on computing the full attention matrix, and hence cannot be used with linear attentions since the full attention weights are not computed explicitly. 

To embed positional information into the XNOR self-attention, or in general for linear transformers that use the kernel trick, candidate encodings must have two characteristics:
\begin{enumerate}
    \item They should depend on relative position $i-j$, i.e., $P(i,j)=f(i-j)$
    \item They should be factorable into positional terms i.e., $P(i,j)=g(i)h(j)$
\end{enumerate}
Using such a definition, the terms $Sm(Q_i)$, $Sm(K_j)$, $Sm'(Q_i)$ and $Sm'(K_j)$ in Equation \ref{equation:softmax_wxnor} can be modified to $\hat{Sm}(Q_i) = Sm(Q_i)g(i)$, $\hat{Sm}(K_j) = Sm(K_j)h(j)$, $\hat{Sm}'(Q_i) = Sm'(Q_i)g(i)$ and $\hat{Sm}'(K_j) = Sm'(K_j)h(j)$ respectively. $\mathcal{O}_i$ can now be computed using the modified values to incorporate position information.


Based on prior work, we adopt two formulations for the positional encoding: a cosine form \cite{zhen2022cosformer}, and rotational positional encodings \cite{su2021roformer}. The cosine positional encodings are realized using Ptolemy's theorem as 
\begin{equation}
\label{equation:cos_pos}
\begin{split}
    P(i,j) &= cos\Big(\frac{\pi(i-j)}{2M}\Big) \\
    &= cos\Big(\frac{\pi i}{2M}\Big)cos\Big(\frac{\pi j}{2M}\Big)+sin\Big(\frac{\pi i}{2M}\Big)sin\Big(\frac{\pi j}{2M}\Big)
\end{split}
\end{equation}
Here $M$ is the maximum sequence length in the batch of examples to ensure that the the $cos(x)$ is computed within $0 < x < \pi/2$

Rotational positional encoding encode absolute positional information within rotation matrices. The product of rotation matrices encodes relative information. Equation \ref{equation:rot_pos} defines the rotational positional embedding, where $\mathbb{R}^d_{\Theta,j}$ is a d-dimensional rotation matrix that encodes the absolute position $j$.  We refer interested readers to \cite{su2021roformer} for a complete mathematical treatment.

\begin{equation}
\label{equation:rot_pos}
    P(i,j) = \mathbb{R}^d_{\Theta,n-m} = \mathbb{R}^d_{\Theta,m}\mathbb{R}^d_{\Theta,n}
\end{equation}

\section{Experimental Setup}
\label{sec:setup}

\label{sec:setup}
We evaluate the proposed model on two speech tasks: speech recognition and speech summarization.
Speech signals are typically very long (comprising sequences of thousands or even tens of thousands of vectors), and are well suited to bring out the effectiveness of our proposed solutions.

\subsection{Speech Recognition}
End to End Speech Recognition is a sequence transduction task which maps a sequence of input speech frames to sequences of language tokens. Speech recognition is usually performed on individual utterances, with input sequences that are a few hundred vectors long, and output sequences that are generally less than 100 output tokens.

We use state-of-the-art conformer based sequence models with attention to perform speech recognition. 
%
Conformers \cite{Guo2021ESPConformer} are variants of the standard transformer architecture that contain macaron-style feedforward layers that sandwich a self-attentive and convolutional layer. Our models have a conformer encoder and a standard transformer decoder\cite{vaswani2017}. Our baseline uses Multihead Self-Attention with Relative Positional Encodings\cite{dai-etal-2019-transformer}. The proposed approach replaces these with XNOR and W-XNOR self-attentions as formulated in Section  \ref{sec:xnor_selfattn}.
 
 We perform our experiments on two corpora: Librispeech-100 \cite{vassil2015} and INTERVIEW-300. Librispeech is a public corpus with read speech from audio-books. INTERVIEW is a corpus of two-party radio conversations between a host and guest collected by NPR \cite{majumder-etal-2020-interview, zhu-etal-2021-mediasum}.
 INTERVIEW-300 is a 300 episode subset of the whole corpus which contains around 150h of speech. To the best of our knowledge, ours is the first work to report speech recognition results on the INTERVIEW corpus. Speech Recognition performance is evaluated using Word Error Rate (WER) on the test sets.

\subsubsection{Librispeech-100 Models}
Our model uses a conformer encoder with 4-fold convolutional subsampling followed by 12 encoder layers with feed-forward dimension 1024 and 4 attention heads. The transformer decoder has 6 layers with feed-forward dimension 2048 and 4 attention heads. Three fold speed perturbation is used for training with speeds 0.9x, 1.0x and 1.1x. SpecAug\cite{park2019specaug} is used with time warping, frequency masking, and time masking. We use 80-dimensional filter-bank features extracted at 100 frames/sec with 25 ms windows. We use the Adam optimizer with peak learning rate 0.002, and 15,000 warmup steps. 

The models are trained with joint Connectionist Temporal Classification (CTC)-Attention \cite{kim2017joint} with the weight for CTC training set to 0.3. Inference is performed using output synchronous beam search with CTC weight of 0.3, and beam width 20. 

\subsubsection{INTERVIEW-300 Models}
 The Interview-300 model uses a conformer encoder with 2-fold convolutional subsampling followed by 6 encoder layers with feed-forward dimension 2048, and 8 attention heads. The transformer decoder has 6 layers with feed-forward dimension 2048 and 8 attention heads. HUBERT \cite{DBLP:journals/taslp/HsuBTLSM21} features from a Librivox model are used as inputs to our models. The models are trained with joint CTC-Attention \cite{kim2017joint} with the weight for CTC training set to 0.3. Inference is performed using output synchronous beam search with CTC weight of 0.3, and beam width 20. 



\subsection{Speech Summarization}
Abstractive Speech summarization is the task of directly extracting an abstractive text summary from speech in videos or meetings. How2 \cite{sanabria18how2} is a dataset of instructional YouTube videos, manually annotated for transcript, Portuguese translation and abstractive summary. 
For this task, generally, multiple utterances of speech are used as input. \cite{Sharma2022} finds that the summarization task uses 35x longer input sequences for summarization when compared to speech recognition on the How2 dataset of instructional videos. 

We utilize a similar two-stage training mechanism where utterance level speech recognition is used as supervised pre-training. The model is then fine-tuned on video-level speech for summarization. To reduce compute, we use the 300h subset of the How2 data for pre-training and the 2000h subset for summarization performance.Therefore, our summarization performance is not comparable with previous approaches\cite{Sharma2022}. 

The conformer encoder uses 2-fold convolutional subsampling followed by 6 encoder layers with feed-forward dimension 2048, and 8 attention heads. The transformer decoder has 6 layers with feed-forward dimension 2048 and 8 attention heads. ASR models are trained with joint CTC-Attention \cite{kim2017joint} with the weight for CTC training set to 0.3. The videos are trimmed to 100s for the video-level speech tasks. HUBERT \cite{DBLP:journals/taslp/HsuBTLSM21} features from a Librivox model are used as inputs to our models. 

\section{Experimental Results}
\label{sec:results}
\label{sec:results}

\begin{table}[h]
\caption{WER (\%) on test-other and test-clean sets from the Librispeech evaluation of models trained on Librispeech-100. Higher numbers are better 
}
\begin{tabular}{c|c|c|c|c}
\hline
Model Description                              & Kernel                                                                  & \begin{tabular}[c]{@{}c@{}}Pos. \\ Enc.\end{tabular} & \begin{tabular}[c]{@{}c@{}}Test-clean\\ WER. (\%)\end{tabular} &  \begin{tabular}[c]{@{}c@{}}Test-other\\ WER. (\%)\end{tabular} \\ \hline

Multi-head Attention & None                                                                    & RelPos                                               &                                           6.6                 &   17.3       \\ \hline 
Linear Transformer \cite{katharopoulos_et_al_2020} & ELU +1 & None & 9.2 & 23.6 \\
Cosformer \cite{zhen2022cosformer}                                      & ReLU                                                                   & cos                                         &  10.8                                                          &  26.3        \\ \hline
Softmax Kernel                                 & \multirow{2}{*}{Softmax}                                                & None                                                 &  10.5                                                          & 29.0         \\
                                               &                                                                         & cos                                                  &       9.0                                                     &   23.5       \\ \hline 
XNOR-former & XNOR & None & 8.5 & 21.9 \\
WXNOR-former                                    & \multirow{3}{*}{\begin{tabular}[c]{@{}c@{}}XNOR\\ Softmax\end{tabular}} & None                                                 &       8.7                                                     &    23.0      \\
                                               &                                                 & cos                                                  &       \textbf{8.1}                                                     &  \textbf{21.3}        \\
                                               &                                                                         & RoPE                                                 &    8.7                              & 23.6         \\
                                             \hline      
\end{tabular}
\label{tab:librispeech_100}
\end{table}

\begin{table}[h]
\caption{Results of our models on the speech summarization task for the How2-2000h data. Abstractive Summarization is evaluated using Rouge scores (R-1,R-2,R-L), METEOR(MTR) scores for content, and BERTScore(BERT) for semantic relevance. Higher numbers are better. }
\resizebox{\columnwidth}{!}{%
\begin{tabular}{l|l|l|l|l|l}
\hline
\multicolumn{1}{l|}{Model} & R-1 & R-2  & R-L                   & MTR                & BERTS \\ \hline
W-XNOR                              & \textbf{53.58}    & \textbf{34.23} & \multicolumn{1}{l|}{\textbf{47.92}} & \multicolumn{1}{l|}{\textbf{25.47}} &   \textbf{89.79}        \\
Cosformer                               &  47.91   & 27.85 & \multicolumn{1}{l|}{42.49} & \multicolumn{1}{l|}{21.26} &   89.36       \\ \hline
\end{tabular}
}
\label{tab:speech_summarization}
\end{table}

\begin{table}[h]
\caption{WER (\%) performance on the INTERVIEW-300 dataset. Lower WER is better}
\begin{tabular}{l|l|l}
\hline
\multicolumn{1}{l|}{Self-Attention Type} & \begin{tabular}[c]{@{}l@{}}Valid \\ Acc\end{tabular} & \multicolumn{1}{l}{\begin{tabular}[c]{@{}l@{}}Test\\ WER (\%)\end{tabular}} \\ \hline
Standard MHA with RelPos                  &   89.3                                                   & 15.2                                                                         \\
Cosformer                                 &       87.1                                               & 15.6                                                                         \\
XNORformer                                &    88.0                                                  & 14.8                                                                         \\
WXNORformer                               &   87.5                                                   & \textbf{14.6}    \\ \hline                                                                    
\end{tabular}
\label{tab:interview_300}
\end{table}

\subsection{Speech Recognition}
For Librispeech-100, we begin by training models using standard multi-head attention and then on comparable linear transformers\cite{katharopoulos_et_al_2020,zhen2022cosformer}. From Table \ref{tab:librispeech_100} 
, we observe that there exists a substantial gap between the WERs of the original transformer and linear transformer variants. If we use a softmax kernel with no positional information, and with cosine embeddings as formulated in Equation \ref{equation:cos_pos}, we observe that the cosine embeddings improve performance, and the softmax kernel performs as well as the linear transformer\cite{katharopoulos_et_al_2020}. 

On the other hand, the proposed W-XNOR self-attention with cosine positional embeddings outperforms the best linear transformer by 5 - 10 \% relative on the test-clean and test-other sets. Further, the cosine positional embeddings appear to be more useful than rotational positional embeddings. 

On the INTERVIEW-300 set, we compare the standard multi-head attention with the Cosformer and our proposed XNOR variants. The proposed approach improves speech recognition performance by 1 \% absolute over the state of the art Cosformer, while slightly outperforming the standard multi-head attention model as well.

\subsection{Speech Summarization-How2}

On End-to-End Speech Summarization, Table \ref{tab:speech_summarization} highlights the results. We observe that the proposed W-XNOR self-attention based model outperforms the Cosformer on a very long input sequence (10k) task without extensive hyperparameter tuning. The XNOR model has a better performance in terms of the ROUGE-L, METEOR and BERTScore metrics, which demonstrates that the proposed W-XNOR model produces more coherent, semantically relevant, content rich summaries in comparison to the baseline.


 
\begin{figure}[h]
    \centering
    \includegraphics[width=0.5\textwidth]{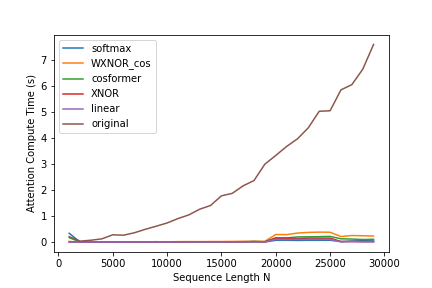}
    \caption{Plot showing compute times for different sequence lengths across linear transformer variants, and full softmax attention}
    \label{fig:scaling}
\end{figure}

\begin{figure}[h]
    \centering
    \includegraphics[width=0.5\textwidth]{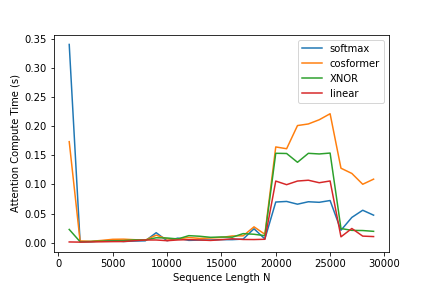}
    \caption{Plot showing compute times for different sequence lengths across linear transformer variants}
    \label{fig:scaling_linear}
\end{figure}

 \subsection{Speed and Scaling}

Figure \ref{fig:scaling} demonstrates the increase in compute time when the sequence length is increased from 1 to 30k over different attention variants. We observe that the linear self-attention variants significantly improve compute time over standard multihead attention. Figure \ref{fig:scaling_linear} presents another view of the same plot with only linear self-attention variants considered. We note that for very long sequences, the softmax kernel is the fastest, followed by ELU\cite{katharopoulos_et_al_2020}, the proposed XNOR self-attention, and then the Cosformer \cite{zhen2022cosformer}. It is clear that our proposed model yields significant performance gains on ASR and speech summarization with speeds significantly faster than standard self-attention, and comparable to other linear transformers. 


\section{Conclusion and Future Work}
\label{sec:conclusion}

In this work, we address the challenge of quadratic computational complexity in transformer self-attentions which makes them intractable over long input sequences. Self-attention is formulated as the softmax over an $N \times N $ matrix, and can be linearized using the kernel trick and associativity of matrix multiplication. 

However there exists a significant gap in performance between full softmax attention and linear transformers. We propose develop the \emph{XNOR self-attention}, and correspondingly the \emph{WXNOR transformer}. The proposed approach outperforms previous linear transformers on the speech recognition and summarization benchmarks.
We also demonstrate that the  performance improvements are achieved while maintaining slightly better or comparable compute efficiency. 

\bibliographystyle{IEEEbib}
\bibliography{strings,refs}

\end{document}